\documentclass{article}
\usepackage[margin=2.5cm]{geometry}

\usepackage{amsmath,xfrac,stackengine,amsthm}
\usepackage{graphicx}
\usepackage{subfigure} 
\usepackage{amssymb}
\usepackage{dsfont} 
\usepackage{xcolor}
\usepackage[font=small,skip=2pt]{caption}
\usepackage[english]{babel}
\usepackage{comment}
\usepackage{enumitem}

\newcommand{\p}[1]{\widetilde{#1}}
\renewcommand{\d}[1]{\delta #1}

\newcommand{\sign}[1]{\text{sgn}(#1)}
\newcommand{\trace}[1]{\text{tr}(#1)}
\newcommand{\spanning}[1]{\text{span}(#1)}
\newcommand{\norm}[1]{\left\lVert#1\right\rVert}
\renewcommand{\exp}[1]{\text{exp}\hspace{-2px}\left(#1\right)}
\newcommand{\snorm}[1]{\left\lVert#1\right\rVert_{\psi_2}}
\newcommand{\svnorm}[1]{\left\lVert#1\right\rVert_{\Psi_2}}

\newcommand{\Prob}[1]{{\mathbf{P}\hspace{-1px}\left(#1\right)}}
\newcommand{\E}[1]{\mathbf{E}\hspace{-1px}\left[#1\right]}
\newcommand{\Var}[1]{\mathbf{Var}\hspace{-1px}\left[#1\right]}

\newcommand{\inner}[2]{\langle #1, #2\rangle}
\providecommand{\abs}[1]{\left\lvert #1 \right\rvert}

\renewcommand{\paragraph}[1]{\vspace{2mm}\noindent\textbf{#1}}

\newtheoremstyle{style}
{6pt} 
{2pt} 
{\itshape} 
{} 
{\bfseries} 
{.} 
{.5em} 
{} 

\theoremstyle{style}
\newtheorem{theorem}{Theorem}[section]

\newtheorem{lemma}{Lemma}[section]

\newtheorem{corollary}{Corollary}[section]

\newtheorem{question}{Problem}

\title{How close are the eigenvectors and eigenvalues of the sample and actual covariance matrices?}
\author{Andreas Loukas \\
\'{E}cole Polytechnique F\'{e}d\'{e}rale Lausanne, Switzerland}

\begin{document} 
\maketitle

\begin{abstract}
How many samples are sufficient to guarantee that the eigenvectors and eigenvalues of the sample covariance matrix are close to those of the actual covariance matrix? 
For a wide family of distributions, including distributions with finite second moment and distributions supported in a centered Euclidean ball, we prove that the inner product between eigenvectors of the sample and actual covariance matrices decreases proportionally to the respective eigenvalue distance. 
Our findings imply \textit{non-asymptotic} concentration bounds for eigenvectors, eigenspaces, and eigenvalues. They also provide conditions for distinguishing principal components based on a constant number of samples. 
\end{abstract}

\section{Introduction}

The covariance matrix $C$ of an $n$-dimensional distribution is an integral part of data analysis, with numerous occurrences in machine learning and signal processing. 
It is therefore crucial to understand how close it is to the \textit{sample covariance}, i.e., the matrix $\p{C}$ estimated from a finite number of samples $m$. 
Following developments in the tools for the concentration of measure, Vershynin showed that a sample size of $m = O(n)$ is up to iterated logarithmic factors sufficient for all distributions with finite fourth moment supported in a centered Euclidean ball of radius $O(\sqrt{n})$~\cite{vershynin2012close}. Similar results hold also for sub-exponential distributions~\cite{adamczak2010quantitative} and distributions with finite second moment~\cite{rudelson1999random}.

We take an alternative standpoint and ask if we can do better when only a subset of the spectrum is of interest. 
Concretely, our objective is to characterize how many samples are sufficient to guarantee that an eigenvector and/or eigenvalue of the sample and actual covariance matrices are, respectively, sufficiently close. 
Our approach is motivated by the observation that methods that utilize the covariance commonly prioritize the estimation of principal eigenspaces. For instance, in (local) principal component analysis we are usually interested in the first few eigenvectors~\cite{berkmann1994computation,kambhatla1997dimension}, whereas when reducing the dimension of a distribution one commonly projects it to the span of the first few eigenvectors~\cite{jolliffe2002principal,frostig2016principal}.

Our finding is that the ``spectral leaking'' occurring in the eigenvector estimation is strongly localized w.r.t. the eigenvalue axis. In other words, the eigenvector $\p{u}_i$ of the sample covariance is less far likely to lie in the span of an eigenvector $u_j$ of the actual covariance when the eigenvalue distance $|\lambda_i - \lambda_j|$ 
is large and the concentration of the distribution in the direction of $u_j$ is small. 
This phenomenon agrees with the intuition that principal components of high variance are easier to estimate, exactly because they are more likely to appear in the samples of the distribution.
In addition, it suggests that it might be possible to obtain good estimates of well separated principal eigenspaces from fewer than $n$ samples.

We provide a mathematical argument confirming this phenomenon. Under fairly general conditions, we prove  that 
\begin{align}
    m = O\Bigg(\frac{k_j^2}{ (\lambda_i - \lambda_j)^2}\Bigg) \quad \text{and} \quad m =  O\Bigg( \frac{k_i^2}{\lambda_i^2}\Bigg) 
\end{align}
samples are asymptotically almost surely (a.a.s). sufficient to guarantee that $|\inner{\p{u}_i}{u_j}|$ and $|\d{\lambda}_i|/\lambda_i$, respectively, is small for all distributions with finite second moment. Here, $k_j^2$ is a measure of the kurtosis of the distribution in the direction of $u_j$. 
We also attain a high probability bound for distributions supported in a centered and scaled Euclidean ball, %
and show how our results can be used to characterize the sensitivity of principal component analysis to a limited sample set. 

\begin{figure*}[t!]
\centering    
\subfigure[$m = 10$]{\label{fig:a}\includegraphics[width=0.245\textwidth]{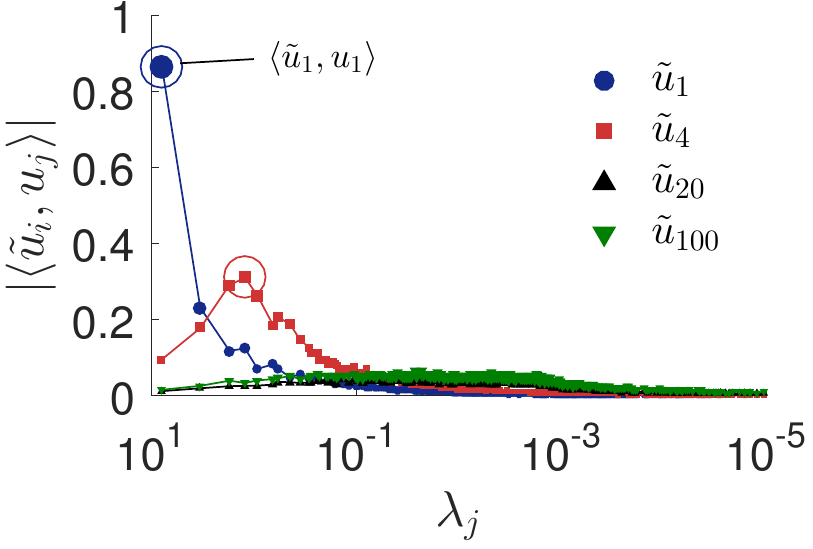}}
\subfigure[$m = 100$]{\label{fig:b}\includegraphics[width=0.245\textwidth]{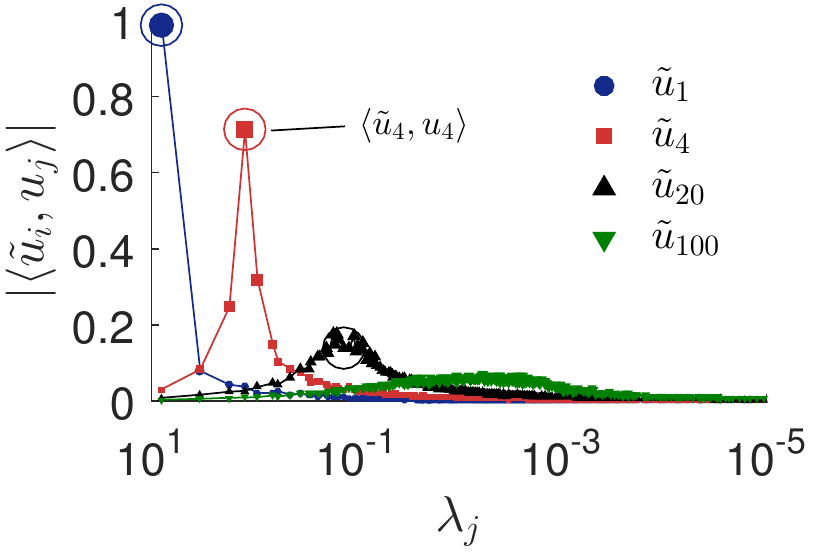}}
\subfigure[$m = 500$]{\label{fig:b}\includegraphics[width=0.245\textwidth]{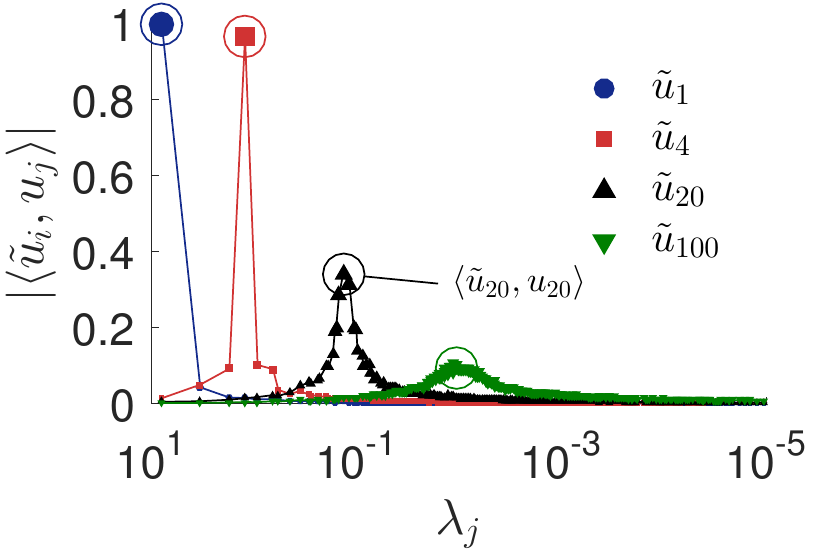}}
\subfigure[$m = 1000$]{\label{fig:b}\includegraphics[width=0.245\textwidth]{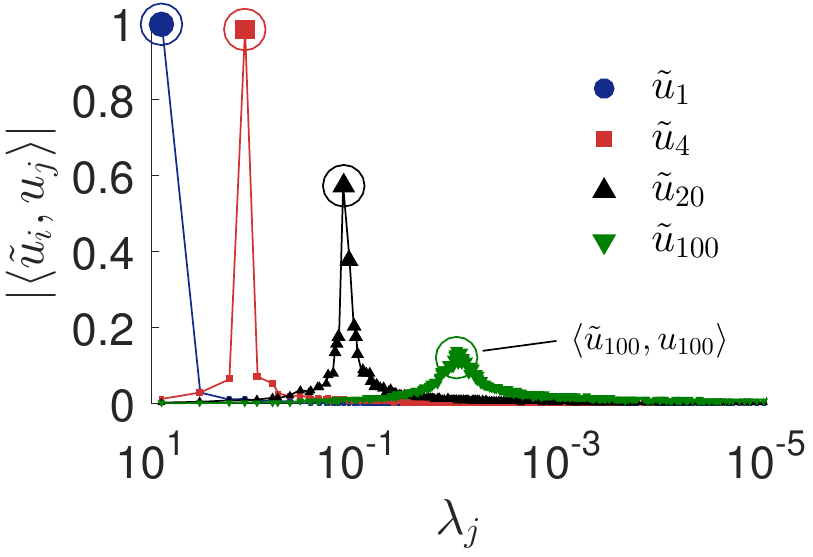}}
\caption{Inner products $\inner{\p{u}_i}{u_j}$ are localized w.r.t. the eigenvalue axis. The phenomenon is shown for MNIST. 
Much fewer than $n = 784$ samples are needed to estimate $u_1$ and $u_4$. }
\label{fig:localized}
\end{figure*}

To the best of our knowledge, these are the first non-asymptotic results concerning the eigenvectors of the sample covariance of distributions with finite second moment. 
Previous studies have intensively investigated the limiting distribution of the eigenvalues of a sample covariance matrix~\cite{silverstein1995empirical,bai1999methodologies}, such as the smallest and largest eigenvalues~\cite{bai1993limit} and the eigenvalue support~\cite{bai1998no}.
Eigenvectors and eigenprojections have attracted less attention; the main research thrust entails using tools from the theory of large-dimensional matrices to characterize limiting distributions~\cite{anderson1963asymptotic,girko1996strong,schott1997asymptotics,bai2007asymptotics} and it has limited applicability in the non-asymptotic setting where the sample size $m$ is small and $n$ cannot be arbitrary large.

Differently, our arguments follow from a combination of techniques from perturbation analysis and non-asymptotic concentration of measure. 
Moreover, in contrast to standard perturbation bounds~\cite{davis1970rotation,yu2015useful} commonly used to reason about eigenspaces~\cite{huang2009spectral,hunter2010}, they can be used to characterize weighted linear combinations of $\inner{\p{u}_i}{u_j}^2$ over $i$ and $j$, and they do not depend on the minimal eigenvalue gap separating two eigenspaces but rather on all eigenvalue differences. The latter renders them particularly amendable to situations where the eigenvalue gap is not significant but the eigenvalue magnitudes decrease sufficiently fast. %

Our work also connects to subspace methods, where the signal and noise spaces are separated by an appropriate eigenspace projection. In their recent work, Shaghaghi and Vorobyov characterized the first two moments of the projection error, a result which implies sample estimates~\cite{shaghaghi2015subspace}. Their results are particularly tight, but are restricted to specific projectors and Normal distributions. 
Finally, we remark that there exist alternative estimators for the spectrum of the covariance with better asymptotic properties~\cite{ahmed1998large,mestre2008improved}. Instead, we here focus on the standard estimates, i.e., the eigenvalues and eigenvectors of the sample covariance.

\section{Problem statement and main results}

Let $x \in \mathbb{C}^n$ be a sample of a multivariate distribution and denote by $x_1, x_2, \ldots, x_m$ the $m$ independent samples used to form the sample covariance, defined as
\begin{align}
    \p{C} = \sum_{p = 1}^m \frac{(x_p - \bar{x} )(x_p - \bar{x})^*}{m},    
\end{align}
where $\bar{x}$ is the sample mean.
Denote by $u_i$ the eigenvector of $C$ associated with eigenvalue $\lambda_i$, and correspondingly for the eigenvectors $\p{u}_i$ and eigenvalues $\p{\lambda}_i$ of $\p{C}$, such that $\lambda_1 \geq \lambda_2 \geq \ldots \geq \lambda_n$. We ask:
\begin{question}
    How many samples are sufficient to guarantee that the inner product $|\inner{\p{u}_i}{u_j}| = |\p{u}_i^* u_j|$ and the eigenvalue gap $|\d{\lambda}_i| = |\p{\lambda}_i - \lambda_i|$ is smaller than some constant $t$ with probability larger than $\epsilon$? 
\end{question}

Clearly, when asking that all eigenvectors and eigenvalues of the sample and actual covariance matrices are close, we will require at least as many samples as needed to ensure that $\| \p{C} - C\|_2 \leq t$~\cite{vershynin2012close}. However, we might do better when only a subset of the spectrum is of interest.  
The reason is that inner products $|\inner{\p{u}_i}{u_j}|$ possess strong localized structure along the eigenvalue axis. 
To illustrate this phenomenon, let us consider the distribution constructed by the $n = 784$ pixel values of digit `1' in the MNIST database. Figure~\ref{fig:localized}, compares the eigenvectors $u_j$ of the covariance computed from all 6742 images, to the eigenvectors $\p{u}_i$ of the sample covariance matrices $\p{C}$ computed from a random subset of $m = 10$, 100, 500, and  $1000$ samples. For each $i = 1, 4, 20, 100$, we depict at $\lambda_j$ the average of $|\inner{\p{u}_i}{u_j}|$ over 100 sampling draws. We observe that: (\textit{i}) The magnitude of $\inner{\p{u}_i}{u_j}$ is inversely proportional to their eigenvalue gap $|\lambda_i - \lambda_j|$. (\textit{ii}) Eigenvector $\p{u}_j$ mostly lies in the span of eigenvectors $u_j$ over which the distribution is concentrated.

We formalize these statements in two steps.

\subsection{Perturbation arguments}
First, we work in the setting of Hermitian matrices and notice the following inequality:

\begingroup
\def\thetheorem{\ref{theorem:perturbation_bound2}}
\begin{theorem}
For Hermitian matrices $C$ and $\p{C} = \d{C} + C$, with eigenvectors  $u_j$ and $\p{u}_i$ respectively, the inequality
\begin{align}
     \abs{\inner{\p{u}_j}{u_j}} \leq \frac{ 2\, \| \d{C} u_j \|_2 }{ \abs{\lambda_i  - \lambda_j} }, \notag
\end{align}
holds for $ \sign{\lambda_i > \lambda_j} \, 2 \p{\lambda}_i > \sign{\lambda_i > \lambda_j} (\lambda_i + \lambda_j)$ and $ \lambda_i \neq \lambda_j$.
\end{theorem}
\addtocounter{theorem}{-1}
\endgroup

The above stands out from standard eigenspace perturbation results, such as the sin($\Theta$) Theorem~\cite{davis1970rotation} and its variants~\cite{huang2009spectral,hunter2010,yu2015useful} for three main reasons:
First, Theorem~\ref{theorem:perturbation_bound2} characterizes the angle between \emph{any} pair of eigenvectors allowing us to jointly bound any linear combination of inner-products. Though this often proves handy (c.f. Section~\ref{sec:application}), it was not possible using sin($\Theta$)-type arguments.
Second, classical bounds are not appropriate for a probabilistic analysis as they feature ratios of dependent random variables (corresponding to perturbation terms). In the analysis of spectral clustering, this complication was dealt with by assuming that $|{\lambda}_i - \lambda_j| \leq |\p{\lambda}_i - \lambda_j|$~\cite{hunter2010}. We weaken this condition at a cost of a multiplicative factor of 2. In contrast to previous work, we also prove that the condition is met a.a.s.
Third, previous bounds are expressed in terms of the minimal eigenvalue gap between eigenvectors lying in the interior and exterior of the subspace of interest. This is a limiting factor in practice as it renders the results only amenable to situations where there is a very large eigenvalue gap separating the subspaces. The proposed result improves upon this by considering every eigenvalue difference.

\subsection{ Spectral concentration}
The second part of our analysis focuses on the covariance and has a statistical flavor.
In particular, we give an answer to Problem 1 for various families of distributions. 

In the context of \textit{distributions with finite second moment}, we prove in Section~\ref{subsec:general} that:
\begingroup
\def\thetheorem{\ref{theorem:prob_arbitrary}}
\begin{theorem}
For any two eigenvectors $\p{u}_i$ and $u_j$ of the sample and actual covariance respectively, and for any real number $t>0$: 
\begin{align}
    \Prob{ |\inner{\p{u}_i}{u_j}| \geq t } \leq \frac{1}{m}\,\Big(\frac{ 2 k_j}{ t \, |\lambda_i - \lambda_j|}\Big)^2,
\end{align}
s.t. the same conditions as Theorem~\ref{theorem:perturbation_bound2}.
\end{theorem}
\addtocounter{theorem}{-1}
\endgroup
For eigenvalues, we provide the following guarantee:
\begingroup
\def\thetheorem{\ref{theorem:eigenvalue}}
\begin{theorem}
For any eigenvalues $\lambda_i$ and $\p{\lambda}_i$ of $C$ and $\p{C}$, respectively, and for any $t > 0$, we have 
\begin{align}
    \Prob{ \frac{|\p{\lambda}_i - \lambda_i|}{\lambda_i} \geq t} \leq \frac{1}{m} \left(\frac{k_i}{\lambda_i\, t}\right)^2. \notag 
\end{align}
%
\end{theorem}
\addtocounter{theorem}{-1}
\endgroup

Term $k_j = (\E{\|xx^*u_j\|_2^2} - \lambda_j^2)^{1/2}$ captures the tendency of the distribution to fall in the span of $u_j$: the smaller the tail in the direction of $u_j$ the less likely we are going to confuse $\p{u}_i$ with $u_j$. 

For \textit{normal distributions}, we have that $k_j^2 = \lambda_j^2 + \lambda_j\trace{C}$ and the number of samples needed for $|\inner{\p{u}_i}{u_j}|$ to be small is
$m = O({\trace{C}}/{\lambda_i^2})$ when $\lambda_j = O(1)$
and
$m = O(\lambda_i^{-2})$ when $\quad \lambda_j = O(\trace{C}^{-1})$. Thus for normal distributions, principal components $u_i$ and $u_j$ with $\text{min}\{ \lambda_i/\lambda_j, \lambda_i\} = \Omega(\trace{C}^{1/2})$ can be distinguished given a constant number of samples.  
On the other hand, estimating $\lambda_i$ with small relative error requires $m = O({\trace{C}}/{\lambda_i})$ samples and can thus be achieved from very few samples when $\lambda_i$ is large\footnote{Though the same cannot be stated about the absolute error $|\d{\lambda_i}|$, that is smaller for small $\lambda_i$.}. 

In Section~\ref{subsec:bounded}, we also give a sharp bound for the family of distributions supported within a ball (i.e., $\|x\| \leq r$ a.s.). 
\begingroup
\def\thetheorem{\ref{theorem:prob_bounded}}
\begin{theorem}
For sub-gaussian distributions supported within a centered Euclidean ball of radius $r$, there exists an absolute constant $c$, independent of the sample size, such that for any real number $t>0$,  
\begin{align}
    \Prob{ |\inner{\p{u}_i}{u_j}| \geq t } \leq  \exp{ 1 - \frac{c \, m \, \Phi_{ij}(t)^2 }{ \lambda_j \svnorm{x}^2}}, 
\end{align}
where $\Phi_{ij}(t) = \frac{ |\lambda_i - \lambda_j| \, t - 2\lambda_j }{2\, (r^2/\lambda_j - 1)^{1/2}} - 2 \svnorm{x}$
and s.t. the same conditions as Theorem~\ref{theorem:perturbation_bound2}.  
\end{theorem}
\addtocounter{theorem}{-1}
\endgroup
Above, $\svnorm{x}$ is the sub-gaussian norm, for which we usually have $\svnorm{x}=O(1)$~\cite{vershynin2010introduction}. As such, the theorem implies that, whenever $\lambda_i \gg \lambda_j = O(1)$, the sample requirement is with high probability $m = O(r^2/{\lambda_i^2})$.

These theorems solidify our experimental findings shown in Figure~\ref{fig:localized}. Moreover, combined with the analysis of principal component estimation given in Section~\ref{sec:application}, they provide a concrete characterization of the relation between the spectrum (eigenvectors and eigenvalues) of the sample and actual covariance matrix as a function of the number of samples, the eigenvalue gap, and the distribution properties.

\section{Perturbation arguments}
\label{sec:perturbation}

Before focusing on the sample covariance matrix, it helps to study $\inner{\p{u}_i}{u_j}$ in the setting of Hermitian matrices.
The presentation of the results is split in three parts. Section~\ref{subsec:perturbation_equalities} starts by studying some basic properties of inner products of the form $\inner{\p{u}_i}{u_j}$, for any $i$ and $j$. The results are used in Section~\ref{subsec:perturbation_inequality_1} to provide a first bound on the angle between two eigenvectors, and refined in Section~\ref{subsec:perturbation_inequalities}. 

\subsection{Basic observations}
\label{subsec:perturbation_equalities}

We start by noticing an exact relation between the angle of a perturbed eigenvector and the actual eigenvectors of $C$. 

\begin{lemma}
For every $i$ and $j$ in $1, 2, \ldots, n$, the relation $(\p{\lambda}_i  - \lambda_j) \, (\p{u}_i^* u_j) = \sum_{\ell = 1}^n (\p{u}_i^* u_\ell) \, ( u_j^* \d{C} u_\ell)$ holds .
\label{lemma:perturbation_equality}
\end{lemma}

\vspace{-3mm}\begin{proof}
The proof follows from a modification of a standard argument in perturbation theory.
We start from the definition $\p{C} \, \p{u}_i = \p{\lambda}_i \, \p{u}_i$ and write 
\begin{align}
     (C + \d{C}) \, (u_i + \d{u}_i) = (\lambda_i + \d{\lambda}_i) \, (u_i + \d{u}_i).
\end{align}
Expanded, the above expression becomes
\begin{align}
     C\d{u}_i + \d{C} u_i + \d{C} \d{u}_i = \lambda_i\d{u}_i  + \d{\lambda}_i u_i + \d{\lambda}_i\d{u}_i,
           \label{eq:lemmaeq_2}
\end{align}
where we used the fact that $C u_i = \lambda_i u_i$ to eliminate two terms. To proceed, we substitute $\d{u}_i = \sum_{j = 1}^n \beta_{ij} u_\ell$, where $ \beta_{ij} = \d{u}_i^*u_\ell$, into ~\eqref{eq:lemmaeq_2} and multiply from the left by $u_j^*$, resulting to: 
\begin{align}
     \sum_{\ell =1}^n \beta_{ij} u_j^* C u_\ell + u_j^* \d{C} u_i + \sum_{\ell =1}^n \beta_{ij} u_j^*\d{C} u_\ell  = \lambda_i \sum_{\ell =1}^n \beta_{ij} u_j^* u_\ell  + \d{\lambda}_i u_j^* u_i + \d{\lambda}_i \sum_{\ell =1}^n \beta_{ij} u_j^* u_\ell
\end{align}
Cancelling the unnecessary terms and rearranging, we have
\begin{align}
      \d{\lambda}_i u_j^* u_i + (\lambda_i + \d{\lambda}_i - \lambda_j) \beta_{ik}  = u_j^* \d{C} u_i + \sum_{\ell =1}^n \beta_{ij} u_j^* \d{C} u_\ell.
      \label{eq:lemmaeq_1}
\end{align}
At this point, we note that $(\lambda_i + \d{\lambda}_i - \lambda_j) = \p{\lambda}_i  - \lambda_j$ and furthermore that $\beta_{ik} = \p{u}_i^* u_j - u_i^* u_j$. 
With this in place, equation~\eqref{eq:lemmaeq_1} becomes
\begin{align}
      \d{\lambda}_i u_j^* u_i + (\p{\lambda}_i  - \lambda_j) \, (\p{u}_i^* u_j - u_i^* u_j) = u_j^* \d{C} u_i + \sum_{\ell =1}^n (\p{u}_i^*u_\ell)\, u_j^* \d{C} u_\ell - u_j^* \d{C} u_i. 
\end{align}
The proof completes by noticing that, in the left hand side, all terms other than $(\p{\lambda}_i  - \lambda_j) \, \p{u}_i^* u_j$ fall-off, either due to $u_i^* u_j = 0$, when $i \neq k$, or because $\d{\lambda}_i = \p{\lambda}_i  - \lambda_j$, o.w. 
\end{proof}

As the expression reveals, $\inner{\p{u}_i}{ u_j}$ depends on the orientation of $\p{u}_i$ with respect to all other $u_\ell$. 
Moreover, the angles between eigenvectors depend not only on the minimal gap between the subspace of interest and its complement space (as in the sin($\Theta$) theorem), but on every difference $\p{\lambda}_i -\lambda_j$. This is a crucial ingredient to a tight bound, that will be retained throughout our analysis. 
 
\subsection{Bounding arbitrary angles}
\label{subsec:perturbation_inequality_1}

We proceed to decouple the inner products. 
\begin{theorem}
For any Hermitian matrices $C$ and $\p{C} = \d{C} + C$, with eigenvectors  $u_j$ and $\p{u}_i$ respectively, we have that $|\p{\lambda}_i  - \lambda_j|\abs{\inner{\p{u}_i}{u_j}} \leq \| \d{C} \, u_j \|_2$.
\label{theorem:perturbation_bound1}
\end{theorem}
%

\vspace{-3mm}\begin{proof}
We rewrite Lemma~\ref{lemma:perturbation_equality} as 
\begin{align}
      (\p{\lambda}_i  - \lambda_j)^2 (\p{u}_i^* u_j)^2 = \left(\sum\limits_{\ell = 1}^n (\p{u}_i^* u_\ell) \, ( u_j^* \d{C} u_\ell) \right)^2.
\end{align}
We now use the Cauchy-Schwartz inequality 
\begin{align}
      (\p{\lambda}_i  - \lambda_j)^2 (\p{u}_i^* u_j)^2 &\leq \sum\limits_{\ell = 1}^n (\p{u}_i^* u_\ell)^2 \, \sum\limits_{\ell = 1}^n ( u_j^* \d{C} u_\ell)^2 = \sum\limits_{\ell = 1}^n ( u_j^* \d{C} u_\ell)^2 = \| \d{C} \, u_j \|_2^2,
\end{align}
where in the last step we exploited Lemma~\ref{lemma:perturbation_dC}. The proof concludes by taking a square root at both sides of the inequality. 
\end{proof}

\begin{lemma}
$\sum\limits_{\ell = 1}^n ( u_j^* \d{C} u_\ell)^2 =  \| \d{C} \, u_j \|_2^2 $.
\label{lemma:perturbation_dC}
\end{lemma}
%
\vspace{-4mm}\begin{proof}
We first notice that $u_j^* \d{C} u_\ell$ is a scalar and equal to its transpose. Moreover, $\d{C}$ is Hermitian as the difference of two Hermitian matrices. We therefore have that
\begin{align}
      \sum\limits_{\ell = 1}^n ( u_j^* \d{C} u_\ell)^2 
      &= \sum\limits_{\ell = 1}^n u_j^* \d{C} u_\ell  u_\ell^* \d{C} u_j =  u_j^* \d{C} \sum\limits_{\ell = 1}^n (u_\ell  u_\ell^*) \d{C} u_j =  u_j^* \d{C} \d{C} u_j  = \| \d{C} u_j \|_2^2, \notag
\end{align}
matching our claim.
\end{proof}

\subsection{Refinement}
\label{subsec:perturbation_inequalities}

As a last step, we move all perturbation terms to the numerator, at the expense of a multiplicative constant factor.  

\begin{theorem}
For Hermitian matrices $C$ and $\p{C} = \d{C} + C$, with eigenvectors  $u_j$ and $\p{u}_i$ respectively, the inequality
\begin{align}
     \abs{\inner{\p{u}_j}{u_j}} \leq \frac{ 2\, \| \d{C} u_j \|_2 }{ \abs{\lambda_i  - \lambda_j} }, \notag
\end{align}
holds for $ \sign{\lambda_i > \lambda_j} \, 2 \p{\lambda}_i > \sign{\lambda_i > \lambda_j} (\lambda_i + \lambda_j)$ and $ \lambda_i \neq \lambda_j$.
\label{theorem:perturbation_bound2}
\end{theorem}
%
\vspace{-3mm}\begin{proof}
Adding and subtracting $ \lambda_i $ from the left side of the expression in Lemma~\ref{lemma:perturbation_equality} gives  
\begin{align}
      (\d{\lambda}_i + \lambda_i - \lambda_j) \, (\p{u}_i^* u_j) = \sum_{\ell = 1}^n (\p{u}_i^* u_\ell) \, ( u_j^* \d{C} u_\ell).
\end{align}
For $\lambda_i \neq \lambda_j$, the above expression can be re-written as 
\begin{align}
      \abs{\p{u}_i^* u_j} &= \frac{ \abs{\sum\limits_{\ell = 1}^n (\p{u}_i^* u_\ell) \, ( u_j^* \d{C} u_\ell) - \d{\lambda}_i\, (\p{u}_i^* u_j)} }{ \abs{\lambda_i  - \lambda_j} } \leq 2 \max \left \{ \frac{ \abs{\sum\limits_{\ell = 1}^n (\p{u}_i^* u_\ell) \, ( u_j^* \d{C} u_\ell)} }{ \abs{\lambda_i  - \lambda_j} }, \frac{ \abs{ \d{\lambda}_i}\, \abs{\p{u}_i^* u_j} }{ \abs{\lambda_i  - \lambda_j} }\right \}.
      \label{eq:perturbation_eq3}
\end{align}

Let us examine the right-hand side inequality carefully. 
Obviously, when the condition $\abs{\lambda_i  - \lambda_j} \leq 2\abs{ \d{\lambda}_i}$ is not met, the right clause of~\eqref{eq:perturbation_eq3} is irrelevant. 
Therefore, for $ \abs{\d{\lambda}_i} < \abs{\lambda_i - \lambda_j}/2$ the bound simplifies to
\begin{align}
      \abs{\p{u}_i^* u_j} &\leq \frac{ 2 \abs{\sum\limits_{\ell = 1}^n (\p{u}_i^* u_\ell) \, ( u_j^* \d{C} u_\ell)} }{ \abs{\lambda_i  - \lambda_j} }.
\end{align}
Similar to the proof of Theorem~\ref{theorem:perturbation_bound1}, applying the Cauchy-Schwartz inequality we have that
\begin{align}
      \abs{\p{u}_i^* u_j} &\leq \frac{ 2 \, \sqrt{\sum\limits_{\ell = 1}^n (\p{u}_i^* u_\ell)^2 \sum\limits_{\ell = 1}^n ( u_j^* \d{C} u_\ell)^2} }{ \abs{\lambda_i  - \lambda_j} } 
      =  \frac{ 2 \, \| \d{C} u_j \|_2 }{ \abs{\lambda_i  - \lambda_j} }, 
\end{align}
where in the last step we used Lemma~\ref{lemma:perturbation_dC}. To finish the proof we notice that, due to Theorem~\ref{theorem:perturbation_bound2}, whenever $|\lambda_i - \lambda_j| \leq |\p{\lambda}_i - \lambda_j|$, one has 
\begin{align}
      \abs{\p{u}_i^*u_j} \leq \frac{ \| \d{C} \, u_j \|_2 }{ |\p{\lambda}_i  - \lambda_j| } \leq \frac{ \| \d{C} \, u_j \|_2 }{ |\lambda_i  - \lambda_j| } < \frac{ 2 \, \| \d{C} u_j \|_2 }{ \abs{\lambda_i  - \lambda_j} }.
\end{align}
Our bound therefore holds for the union of intervals $\abs{\d{\lambda}_i} < \abs{\lambda_i - \lambda_j}/2$ and $|\lambda_i - \lambda_j| \leq |\p{\lambda}_i - \lambda_j|$, i.e., for $\p{\lambda}_i > (\lambda_i + \lambda_j)/2$ when $\lambda_i > \lambda_j$ and for $\p{\lambda}_i < (\lambda_i + \lambda_j)/2$ when $\lambda_i < \lambda_j$.
\end{proof}

\section{Spectral concentration}

This section builds on the perturbation results of Section~\ref{sec:perturbation} to characterize how far any inner product $\inner{\p{u}_i}{u_j}$ and eigenvalue $\p{\lambda}_i$ are from the ideal estimates.

Before proceeding, we remark on some simplifications employed in the following. W.l.o.g., we will assume that the mean $\E{x}$ is zero and the covariance full rank. Though the case of rank-deficient $C$ is easily handled by substituting the inverse with the Moore-Penrose pseudoinverse, we opt to make the exposition in the simpler setting.
In addition, we will assume the perspective of Theorem~\ref{theorem:perturbation_bound2}, for which the inequality $\sign{\lambda_i > \lambda_j} \, 2 \p{\lambda}_i > \sign{\lambda_i > \lambda_j} (\lambda_i + \lambda_j)$ holds. This event occurs a.a.s. when the gap and the sample size are sufficiently large (see Section~\ref{subsubsec:eigenvalue}), but it is convenient to assume that it happens almost surely. Removing this assumption is possible, but is not pursued here as it leads to less elegant and sharp estimates.   

\subsection{Distributions with finite second moment}
\label{subsec:general}

Our first flavor of results is based on a variant of the Tchebichef inequality and holds for any distribution with finite second moment. 

\subsubsection{Concentration of eigenvector angles} 

We start with the concentration of inner-products $|\inner{\p{u}_i}{u_j}|$.

\begin{theorem}
For any two eigenvectors $\p{u}_i$ and $u_j$ of the sample and actual covariance respectively, with $\lambda_i \neq \lambda_j$, and for any real number $t>0$, we have 
%
\begin{align}
    \Prob{ |\inner{\p{u}_i}{u_j}| \geq t } \leq \frac{1}{m}\, \Big(\frac{ 2\, k_j}{t \, |\lambda_i - \lambda_j|}\Big)^2 \notag
\end{align}
for $ \sign{\lambda_i > \lambda_j} \, 2 \p{\lambda}_i > \sign{\lambda_i > \lambda_j} (\lambda_i + \lambda_j)$ and $k_j = \left(\E{ \| x x^* u_j \|_2^2}  - \lambda_j^2\right)^{1/2}$.
\label{theorem:prob_arbitrary}
\end{theorem}
\vspace{-3mm}\begin{proof}
According to a variant of Tchebichef's inequality~\cite{144675}, for any random variable $X$ and for any real numbers $t > 0$ and $\alpha$:
\begin{align}
    \Prob{ | X - \alpha| \geq t } \leq \frac{\Var{X} + (\E{X} - \alpha)^2}{t^2} .
\end{align}
Setting $X = \inner{\p{u}_i}{u_j}$ and $\alpha = 0$, we have
\begin{align}
    \Prob{ | \inner{\p{u}_i}{u_j} | \geq t } &\leq \frac{\Var{\inner{\p{u}_i}{u_j}} + \E{\inner{\p{u}_i}{u_j}}^2}{t^2} = \frac{\E{\inner{\p{u}_i}{u_j}^2}}{t^2} \leq \frac{4\,\E{\|\d{C} u_j \|_2^2}}{t^2 (\lambda_i - \lambda_j)^2},
\end{align}
where the last inequality follows from Theorem~\ref{theorem:perturbation_bound2}.
We continue by expanding $\d{C}$ using the definition of the eigenvalue decomposition and substituting the expectation.
\begin{align}
    \E{\|\d{C} u_j \|_2^2} 
    &= \E{\|\p{C}u_j - \lambda_j u_j \|_2^2} \notag \\
    &= \E{u_j^* (\p{C} - \lambda_j)(\p{C} - \lambda_j) u_j} \notag \\ 
    &= \E{u_j^*\p{C}^2 u_j}  + \lambda_j^2 - 2 \lambda_j u_j^*\E{\p{C}} u_j \notag \\
    &= \E{u_j^*\p{C}^2 u_j}  - \lambda_j^2. 
\end{align}
In addition, 
\begin{align}
    \E{u_j^*\p{C}^2 u_j} &= \sum_{p,q = 1}^m u_j^*\frac{\E{ (x_p x_p^*) (x_q x_q^*) }}{m^2} u_j \notag \\
    &\hspace{-0mm}= \sum_{p \neq q} u_j^*\frac{\E{x_p x_p^*} \E{x_q x_q^*}}{m^2} u_j + \sum_{p = 1}^m u_j^*\frac{\E{ x_p x_p^* x_p x_p^* }}{m^2} u_j \notag \\
    &\hspace{-0mm}= \frac{m(m-1)}{m^2} \lambda_j^2 + \frac{1}{m} u_j^*\E{ x x^* x x^* } u_j \notag \\
    &\hspace{-0mm}= (1 - \frac{1}{m}) \, \lambda_j^2 + \frac{1}{m}u_j^*\E{ x x^* x x^* } u_j
\end{align}
and therefore 
\begin{align}
    \E{\|\d{C} u_j \|_2^2} &= (1 - \frac{1}{m}) \, \lambda_j^2 + \frac{1}{m} u_j^*\E{ x x^* x x^* } u_j - \lambda_j^2 \notag \\
    &\hspace{-0mm}= \frac{ u_j^*\E{ x x^* x x^* } u_j - \lambda_j^2 }{m} = \frac{ \E{ \| x x^* u_j \|_2^2}  - \lambda_j^2 }{m}. \notag 
\end{align}
Putting everything together, the claim follows.
\end{proof}

The following corollary will be very useful when applying our results.

\begin{corollary}
For any weights $w_{ij}$ and real $t>0$:
\vspace{-1mm}
\begin{align}
    \Prob{  \sum_{i\neq j}  w_{ij} \inner{\p{u}_i}{u_j}^2  > t} \leq \sum_{i \neq j}   \frac{ 4 \, w_{ij} \, k_j^2 }{m\,t \,(\lambda_i - \lambda_j)^2},\notag 
\end{align}
\vspace{-1mm}
where $k_j = \left(\E{ \| x x^* u_j \|_2^2}  - \lambda_j^2\right)^{1/2}$ and $w_{ij} \neq 0$ when $\lambda_i \neq \lambda_j$ and $ \sign{\lambda_i > \lambda_j} \, 2 \p{\lambda}_i > \sign{\lambda_i > \lambda_j} (\lambda_i + \lambda_j)$. 
\label{corollary:weighted_sum}
\end{corollary}
\vspace{-3mm}\begin{proof}
We proceed as in the proof of Theorem~\ref{theorem:prob_arbitrary}:
\begin{align}
    \Prob{\hspace{-1px} \Big(\sum_{i\neq j}  w_{ij} \inner{\p{u}_i}{u_j}^2\Big)^{\frac{1}{2}} \hspace{-2px} > t} &\leq \frac{\E{\sum_{i\neq j} w_{ij} \inner{\p{u}_i}{u_j}^2}}{t^2} \leq \frac{4}{t^2} \sum_{i\neq j} w_{ij} \frac{\E{\| \d{C} u_j \|_2^2}}{(\lambda_i - \lambda_j)^2}
\end{align}
The claim follows by computing $\E{\| \d{C} u_j \|_2^2}$ (as before) and squaring both terms within the probability.
\end{proof}

\subsubsection{Eigenvalue concentration} 
\label{subsubsec:eigenvalue}

A slight modification of the same argument can be used to characterize the eigenvalue relative difference.

\begin{theorem}
For any eigenvalues $\lambda_i$ and $\p{\lambda}_i$ of $C$ and $\p{C}$, respectively, and for any $t > 0$, we have 
\begin{align}
    \Prob{ \frac{|\p{\lambda}_i - \lambda_i|}{\lambda_i} \geq t} \leq \frac{1}{m} \left(\frac{k_i}{\lambda_i\, t}\right)^2, \notag 
\end{align}
where $k_i = (\E{\| x x^* u_i\|_2^2} - \lambda_i)^{1/2}$.
\label{theorem:eigenvalue}
\end{theorem}
\vspace{-3mm}\begin{proof}
Directly from the Bauker-Fike theorem~\cite{bauer1960norms} one sees that
\begin{align}
    \abs{\d{\lambda}_i} \leq \| \p{C} u_i - \lambda_i u_i\|_2 &= \| \d{C} u_i \|_2.
\end{align}
The proof is then identical to the one of Theorem~\ref{theorem:prob_arbitrary}.
\end{proof}
%
As such, the probability the main condition of our Theorems holds is at least
\begin{align}
    \Prob{\sign{\lambda_i > \lambda_j} \, 2 \p{\lambda}_i > \sign{\lambda_i > \lambda_j} (\lambda_i + \lambda_j)} 
    \geq \Prob{ |\p{\lambda}_i - \lambda_i| < \frac{|\lambda_i - \lambda_j|}{2}}
    > 1 - \frac{2 k_i^2}{m|\lambda_i - \lambda_j|}.
\end{align}

\subsubsection{The influence of the distribution} 

As seen by the straightforward inequality 
$    \E{\| x x^*u_j\|_2^2} \leq \E{\|x\|_2^4}$,
$k_j$ connects to the kurtosis of the distribution. However, it also captures the tendency of the distribution to fall in the span of $u_j$. 

To see this, we will work with the uncorrelated random vectors $\varepsilon = C^{-1/2} x $, which have zero mean and identity covariance.
\begin{align}
    k_j^2 &= \E{ u_j^* C^{1/2} \varepsilon \varepsilon^* C \varepsilon \varepsilon^* C^{1/2} u_j } - \lambda_j^2 \notag \\
        &= \lambda_j \, \E{ u_j^* \varepsilon \varepsilon^* C \varepsilon \varepsilon^* u_j } \notag - \lambda_j^2 \\
        &= \lambda_j  (\E{ \| \Lambda^{1/2} U^*\varepsilon \varepsilon^* u_j \|_2^2} - \lambda_j).
\end{align}
If we further set $\hat{\varepsilon} = U^* \varepsilon$, we have
\begin{align}
    k_j^2 &= \lambda_j \Big(\sum_{\ell = 1}^n \lambda_\ell \E{\hat{\varepsilon}(\ell)^2 \hat{\varepsilon}(j)^2} - \lambda_j\Big). 
\end{align}
It is therefore easier to untangle the spaces spanned by $\p{u}_i$ and $u_j$ when the variance of the distribution along the latter space is small (the expression is trivially minimized when $\lambda_j \rightarrow 0 $) or when the variance is entirely contained along that space (the expression is also small when $\lambda_i = 0$ for all $i\neq j$). In addition, it can be seen that distributions with fast decaying tails allow for better principal component identification ($\E{\hat{\varepsilon}(j)^4}$ is a measure of kurtosis over the direction of $u_j$).

For the particular case of a Normal distribution, we provide a closed-form expression.
\begin{corollary}
For a Normal distribution, we have $ k_j^2 = \lambda_j \left( \lambda_j + \trace{C} \right)$.
\end{corollary}
\vspace{-3mm}\begin{proof}
For a centered and normal distribution with identity covariance, the choice of basis is arbitrary and the vector $\hat{\varepsilon} = U^* \varepsilon$ is also zero mean with identity covariance. Moreover, 
for every $\ell \neq j$ we can write $\E{\hat{\varepsilon}(\ell)^2  \hat{\varepsilon}(j)^2} = \E{\hat{\varepsilon}(\ell)^2}  \E{\hat{\varepsilon}(j)^2} = 1$. 
This implies that
\vspace{-1mm}
\begin{align}
    \E{ \| x x^* u_j \|_2^2} &= \lambda_j^2 \E{\hat{\varepsilon}(j)^4} + \lambda_j\sum_{\ell\neq j}^n \lambda_\ell = \lambda_j^2 (3-1) + \lambda_j \trace{C} = 2\lambda_j^2 + \lambda_j\trace{C}
\end{align}
and, accordingly, $k_j^2 = \lambda_j \left( \lambda_j + \trace{C} \right)$.
\end{proof}

\subsection{Distributions supported in a Euclidean ball}
\label{subsec:bounded}

Our last result provides a sharper probability estimate for the family of sub-gaussian distributions supported in a centered Euclidean ball of radius $r$, with their $\Psi_2$-norm
\begin{align}
    \svnorm{x} = \sup_{y \in \mathcal{S}^{n-1}} \snorm{\inner{x}{y}},
\end{align}
where $\mathcal{S}^{n-1}$ is the unit sphere and with the $\psi_2$-norm of a random variable $X$ defined as 
\begin{align}
    \snorm{X} = \sup_{p \geq 1} p^{-1/2} \E{ \abs{X}^p }^{{1}/{p}}.   
\end{align}
Our setting is therefore similar to the one used to study covariance estimation~\cite{vershynin2012close}. Due to space constraints, we refer the reader to the excellent review article~\cite{vershynin2010introduction} for an introduction to sub-gaussian distributions as a tool for non-asymptotic analysis of random matrices. 

\begin{theorem}
For sub-gaussian distributions supported within a centered Euclidean ball of radius $r$, there exists an absolute constant $c$, independent of the sample size, such that for any real number $t>0$, %
\begin{align}
    \Prob{ |\inner{\p{u}_i}{u_j}| \geq t } \leq  \exp{ 1 - \frac{c \, m \, \Phi_{ij}(t)^2 }{ \lambda_j \svnorm{x}^2}}, 
\end{align}
where $\Phi_{ij}(t) = \frac{|\lambda_i - \lambda_j| \, t - 2\lambda_j}{2\, (r^2/\lambda_j - 1)^{1/2}} - 2 \svnorm{x}$, $ \lambda_i \neq \lambda_j$ and $ \sign{\lambda_i > \lambda_j} \, 2 \p{\lambda}_i > \sign{\lambda_i > \lambda_j} (\lambda_i + \lambda_j)$.
\label{theorem:prob_bounded}
\end{theorem}

\vspace{-3mm}\begin{proof}
We start from the simple observation that, for every upper bound $B$ of $|\inner{\p{u}_i}{u_j}|$ the relation  
$
\Prob{ |\inner{\p{u}_i}{u_j}| > t } \leq \Prob{ B > t } 
$
holds. To proceed therefore we will construct a bound with a known tail.
As we saw in Sections~\ref{subsec:perturbation_inequalities} and~\ref{subsec:general},
\begin{align}
    |\inner{\p{u}_i}{u_j}| &\leq \frac{2\, \|\d{C} u_j \|_2}{|\lambda_i - \lambda_j|} \notag \\
    &\hspace{-0mm}= \frac{2 \, \norm{ (1/m) \sum_{p = 1}^m (x_p x_p^*u_j - \lambda_j u_j ) }_2}{ |\lambda_i - \lambda_j| } \notag \\
    &\hspace{-0mm}\leq \frac{2 \, \sum_{p = 1}^m \norm{x_p x_p^*u_j - \lambda_j u_j  }_2}{ m \, |\lambda_i - \lambda_j| } \notag \\
    &\hspace{-0mm}= \frac{2 \, \sum_{p = 1}^m \sqrt{ (u_j^*x_p)^2 (x_p^*x_p) - 2\lambda_j (u_j^*x_p)^2 + \lambda_j^2 } }{ m \, |\lambda_i - \lambda_j| } \notag \\
    &\hspace{-0mm}= \frac{2 \, \sum_{p = 1}^m \sqrt{ (u_j^*x_p)^2 ( \| x_p\|_2^2 - \lambda_j) + \lambda_j^2 } }{ m \, |\lambda_i - \lambda_j| }
\end{align}
Assuming further that $ \|x\|_2 \leq r$, and since the numerator is minimized when $\| x_p\|_2^2$ approaches $\lambda_j$, we can write for every sample $x = C^{1/2} \varepsilon$:
\begin{align}
    \sqrt{ (u_j^*x)^2 ( \| x\|_2^2 - \lambda_j) + \lambda_j^2 } &\leq \sqrt{ (u_j^*x)^2 ( r^2 - \lambda_j) + \lambda_j^2 } \notag \\
    &\hspace{-0mm}=  \sqrt{ \lambda_j (u_j^*\varepsilon)^2 ( r^2 - \lambda_j) + \lambda_j^2 } \leq |u_j^*\varepsilon| \, \sqrt{ \lambda_jr^2 - \lambda_j^2} + \lambda_j,
\end{align}
which is a shifted and scaled version of the random variable $ |\hat{\varepsilon}(j)| = |u_j^*\varepsilon|$. 
Setting $a = ( \lambda_j r^2 - \lambda_j^2)^{1/2}$, we have
\begin{align}
    \Prob{ |\inner{\p{u}_i}{u_j}| \geq t } 
    &\leq \Prob{ \frac{2 \, \sum_{p = 1}^m (|\hat{\varepsilon}_p(j)|\,a + \lambda_j)}{ m \, |\lambda_i - \lambda_j| } \geq t } \notag\\
    &\hspace{-0mm}= \Prob{ \sum_{p = 1}^m (|\hat{\varepsilon}_p(j)|\,a + \lambda_j) \geq 0.5 \, m t \, |\lambda_i - \lambda_j| } \notag \\
    &\hspace{-0mm}= \Prob{ \sum_{p = 1}^m |\hat{\varepsilon}_p(j)| \geq \frac{m \, (0.5 \, t \, |\lambda_i - \lambda_j| - \lambda_j)}{a} }.
    \label{eq:probability_ineq1}
\end{align}
By Lemma~\ref{lemma:probability_subgaussian} however, the left hand side is a sum of independent sub-gaussian variables. Since the summands are not centered, we expand each $|\hat{\varepsilon}_p(j)| = z_p + \E{ |\hat{\varepsilon}_p(j)|}$ in terms of a centered sub-gaussian $z_p$ with the same $\psi_2$-norm. Furthermore, by Jensen's inequality and Lemma~\ref{lemma:probability_subgaussian}
\begin{align}
    \E{ |\hat{\varepsilon}_p(j)| } \leq \E{ \hat{\varepsilon}_p(j)^2 }^{1/2} \leq \frac{2}{\lambda_j} \svnorm{x}.
\end{align}
Therefore, if we set $\Phi_{ij}(t) = \frac{(0.5 \, |\lambda_i - \lambda_j|\, t - \lambda_j)}{( r^2/\lambda_j - 1)^{1/2}} - 2 \svnorm{x}$  
\begin{align}
    \Prob{ |\inner{\p{u}_i}{u_j}| \geq t } 
    &\leq \Prob{ \sum_{p = 1}^m z_p \geq \frac{m \Phi_{ij}(t)}{\lambda_j}}. 
\end{align}
Moreover, by the rotation invariance principle, the left hand side of the last inequality is a sub-gaussian with $\psi_2$-norm smaller than $(C \sum_{p= 1}^m \snorm{z_p}^2)^{1/2} = (c_1 m)^{1/2} \snorm{z} \leq (C m / \lambda_j)^{1/2} \svnorm{x}$, for some absolute constant $c_1$. As a consequence, there exists an absolute constant $c_2$, such that for each $\theta > 0$:
\begin{align}
    \Prob{ \abs{\sum_{p = 1}^m z_p} \geq \theta } \leq \exp{ 1 - \frac{c_2 \, \theta^2 \lambda_j }{ m \svnorm{x}^2}}. 
\end{align}
Substituting $\theta = {m \, \Phi_{ij}(t)}/{\lambda_j}$, we have 
\begin{align}
    \Prob{ |\inner{\p{u}_i}{u_j}| \geq t } 
    &\leq  \exp{ 1 - \frac{c_2 \, m^2 \, \Phi_{ij}(t)^2 \lambda_j }{ m \lambda_j^2 \svnorm{x}^2}}  =  \exp{ 1 - \frac{c_2 \, m \, \Phi_{ij}(t)^2 }{ \lambda_j \svnorm{x}^2}}, 
\end{align}
which is the desired bound.
\end{proof}

\begin{lemma}
    If $x$ is a sub-gaussian random vector and $\varepsilon = C^{-1/2} x$, then for every $i$, the random variable $ \hat{\varepsilon}(i) = u_i^*\varepsilon$ is also sub-gaussian, with $\snorm{\hat{\varepsilon}(i)} \leq \svnorm{x} / \sqrt{\lambda_i}$. 
    \label{lemma:probability_subgaussian}
\end{lemma}
%
\vspace{-3mm}\begin{proof}
The fact that $\hat{\varepsilon}(i)$ is sub-gaussian follows easily by the definition of a sub-gaussian random vector, according to which for every $y\in \mathbb{R}^n$ the random variable $\inner{x}{y}$ is sub-gaussian. Setting $y = u_i$, the first part is proven. 
For the bound of the norm, notice that 
\begin{align}
    \svnorm{x} \hspace{-2px} &= \hspace{-5px} \sup_{y \in \mathcal{S}^{n-1}} \hspace{-5px} \snorm{\inner{x}{y}} \hspace{-3px}= \hspace{-4px}\sup_{y \in \mathcal{S}^{n-1}} \hspace{-2px}\snorm{ \sum_{j = 1}^n \lambda_j^{1/2} (u_j^*y) (u_j^*\varepsilon) } \geq \snorm{ \sum_{j = 1}^n \lambda_j^{1/2} (u_j^*u_i) \hat{\varepsilon}(j) }= \lambda_i^{1/2} \snorm{ \hat{\varepsilon}(i) },
\end{align}
where, for the last inequality, we set $y = u_i$. 
\end{proof}

\section{Principal component estimation}
\label{sec:application}

\begin{figure}[t!]
\centering    
\subfigure[$\ell = 0$]{\label{fig:2a}\includegraphics[width=0.48\columnwidth]{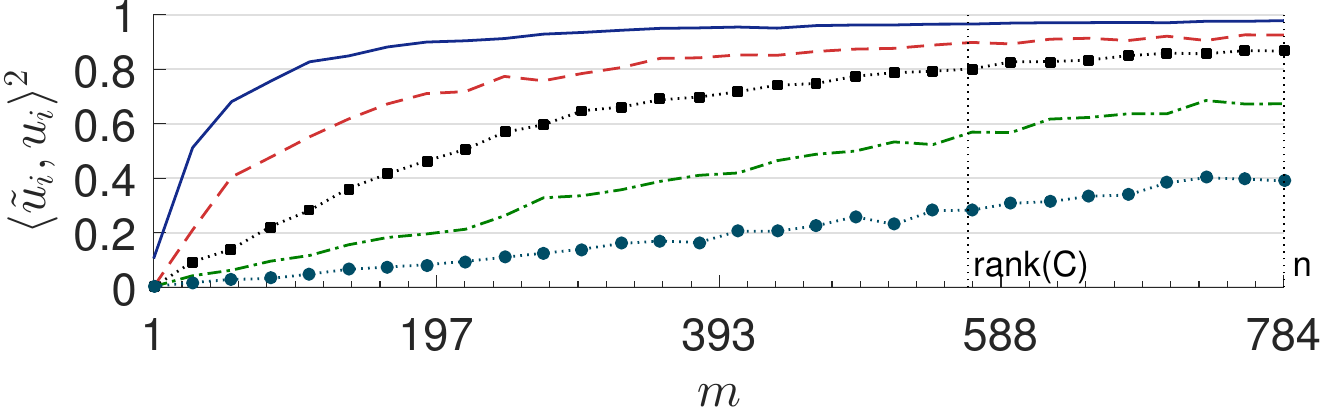}}
\subfigure[$\ell = 2$]{\label{fig:2b}\includegraphics[width=0.48\columnwidth]{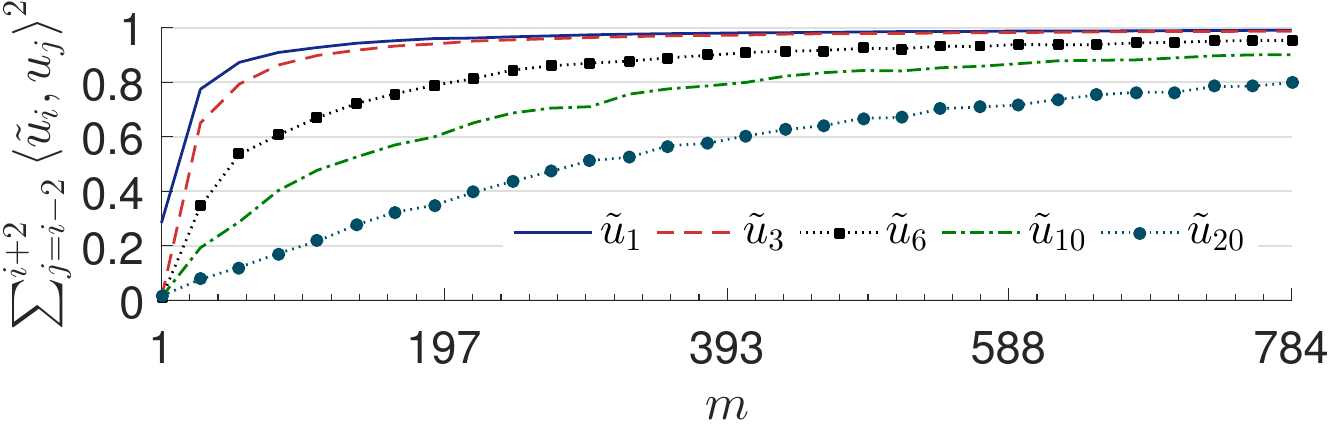}}
\subfigure[$\ell = 4$]{\label{fig:2c}\includegraphics[width=0.48\columnwidth]{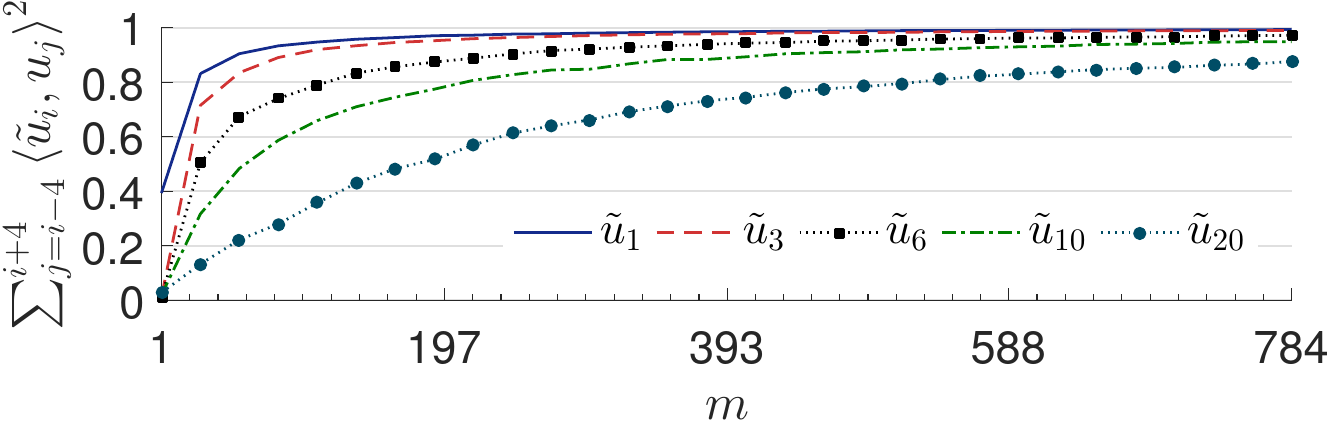}}
\caption{Much less than $n= 784$ samples are needed to estimate principal components associated to large isolated eigenvalues (top). The sample requirement reduces further if we are satisfied by a tight localization of eigenvectors, such that $\p{u}_i$ is almost entirely contained within the span of the $2\ell$ surrounding eigenvectors of $u_i$, shown here for $\ell = 2$ (middle) and $\ell = 4$ (bottom).}
\label{fig:PC}
\end{figure}

The last step in our exposition entails moving from statements about eigenvector angles to statements about the eigenvectors themselves. In particular, we are going to focus on the eigenvectors of the sample covariance associated to the largest eigenvalues, commonly referred to as principal components. 

The phenomenon we will characterize is portrayed in Figure~\ref{fig:PC}. The experiment in question concerns the $n = 784$ dimensional distribution constructed by the 6131 images featuring digit `3' found in the MNIST database. 
The top sub-figure shows the estimation accuracy for a set of five principal components averaged over 200 sampling draws. The trends suggest that the first three eigenvectors can be estimated up to a satisfactory accuracy by much less than $n$ samples. 
The sample requirements decrease further if we are satisfied by a tight localization of eigenvectors\footnote{This is relevant for instance when we wish to construct projectors over specific principal eigenspaces and we have to ensure that the projection space estimated from the sample covariance closely approximates an eigenspace of the actual covariance.}, such that $\p{u}_i$ is almost entirely contained within the span of the $2\ell$ surrounding eigenvectors of $u_i$.  As suggested by the middle and bottom sub-figures, by slightly increasing $\ell$, we reduce the number of samples needed to estimate higher principal components.

In the following, we show that our results verify these trends. We are interested to find out how many samples are sufficient to ensure that $\p{u}_i$ is almost entirely contained within $\spanning{u_{i-\ell}, \ldots, u_{i}, \ldots, u_{i+\ell}}$ for some small non-negative integer $\ell$.
Setting $\mathcal{S} = \{ (j < i - \ell) \cup  (i+\ell < j \leq r) \}$, where $r$ is the rank of $C$, we have as a consequence of Corollary~\ref{corollary:weighted_sum} that for distributions with finite second moment:
\begin{align}
    \Prob{ \sum_{j = i-\ell}^{i+\ell} \hspace{-2px}\inner{\p{u}_i}{u_j}^2 < 1-t} \leq \sum_{j \in \mathcal{S}} \frac{4\,k_j^2}{m t\, (\lambda_i - \lambda_j)^2}
    \label{eq:PC}
\end{align}

In accordance with the experimental results, equation~\eqref{eq:PC} reveals that it is much easier to estimate the principal components of larger variance, and that, by introducing a small slack in terms of $\ell$ one mitigates the requirement for eigenvalue separation.

It might be also interesting to observe that, for covariance matrices that are (approximately) low-rank, we obtain estimates reminiscent of compressed setting~\cite{Candes:2011:RPC:1970392.1970395}, in the sense that the sample requirement becomes a function of the non-zero eigenvalues. Though intuitive, this dependency of the estimation accuracy on the rank was not transparent in known results for covariance estimation~\cite{rudelson1999random,adamczak2010quantitative,vershynin2012close}. 

\section{Conclusions}

The main contribution of this paper was the derivation of non-asymptotic bounds for the concentration of inner-products $|\inner{\p{u}_i}{u_j}|$ involving eigenvectors of the sample and actual covariance matrices. We also showed how these results can be extended to reason about eigenvectors, eigenspaces, and eigenvalues.

We have identified two interesting directions for further research. 
The first has to do with obtaining tighter concentration estimates. Especially with regards to our perturbation arguments, we believe that our current bounds on inner products could be sharpened by at least a constant multiplicative factor. We also suspect that a joint analysis of angles could also lead to a significant improvement over Corollary~\ref{corollary:weighted_sum}.    
The second direction involves using our results for the analysis of methods that utilize the eigenvectors of the covariance for dimensionality reduction. Examples include (fast) principal component projection~\cite{frostig2016principal} and regression~\cite{jolliffe1982note}.

\bibliographystyle{plain}
\bibliography{references}

\end{document}